\RequirePackage{fix-cm}
\documentclass[smallextended]{svjour3}       
\smartqed  
\usepackage{graphicx}
\usepackage{float}
\usepackage[justification=centering]{caption}

\usepackage{amsmath}
\usepackage{amsthm}
\usepackage{amsfonts}
\usepackage{subfigure}
\usepackage{enumitem}
\usepackage{algorithm}
\usepackage{algpseudocode}
\usepackage{graphicx}
\usepackage{booktabs}
\usepackage{natbib}
\usepackage{apalike}
\usepackage[breaklinks]{hyperref}
\usepackage[hyphenbreaks]{breakurl}
\usepackage[figuresright]{rotating}
\usepackage{adjustbox}
\hypersetup{colorlinks=true,citecolor=blue,linkcolor=blue,urlcolor=blue}
\begin{document}

\title{Memory-Efficient Factorization Machines via Binarizing both Data and Model Coefficients
}

\titlerunning{ }        

\author{Yu Geng        \and
        Liang Lan 
}


\institute{Y. Geng and L. Lan \at
              Department of Computer Science, Hong Kong Baptist University, Hong Kong SAR, China \\
              \email{\{yugeng, lanliang\}@comp.hkbu.edu.hk} 
}


\maketitle

\begin{abstract}
Factorization Machines (FM), a general predictor that can efficiently model feature interactions in linear time, was primarily proposed for collaborative recommendation and have been broadly used for regression, classification and ranking tasks. Subspace Encoding Factorization Machine (SEFM) has been proposed recently to overcome the expressiveness limitation of Factorization Machines (FM) by applying explicit nonlinear feature mapping for both individual features and feature interactions through one-hot encoding to each input feature. Despite the effectiveness of SEFM, it increases the memory cost of FM by $b$ times, where $b$ is the number of bins when applying one-hot encoding on each input feature. To reduce the memory cost of SEFM, we propose a new method called Binarized FM which constraints the model parameters to be binary values (i.e., 1 or $-1$). Then each parameter value can be efficiently stored in one bit. Our proposed method can significantly reduce the memory cost of SEFM model. In addition, we propose a new algorithm to effectively and efficiently learn proposed FM with binary constraints using Straight Through Estimator (STE) with Adaptive Gradient Descent (Adagrad). Finally, we evaluate the performance of our proposed method on eight different classification datasets. Our experimental results have demonstrated that our proposed method achieves comparable accuracy with SEFM but with much less memory cost. 
\end{abstract}

\section{Introduction} \label{intro}
Factorization Machines (FM)~\citep{rendle2010factorization} is a popular method that can efficiently model feature interactions in linear time with respect to the number of features. While FM has been successfully applied in different predictive tasks, it is also be noted that FM fails to capture highly nonlinear patterns in the data because it only uses polynomial feature expansion for nonlinear relationship modeling~\citep{he2017neural,liu2017locally,lan2019accurate,chen2019generalized}. To address the expressiveness limitation of the FM model, different methods have been proposed: (1) Embedding neural networks into FM models~\citep{he2017neural,xiao2017attentional,guo2017deepfm}: they use neural-network based approach to model feature interactions in a highly nonlinear way; (2) Locally Linear FM (LLFM)~\citep{liu2017locally} and its extension~\citep{chen2019generalized}: they use a local encoding technique to capture nonlinear concept based on the idea that nonlinear manifold behaves linearly in its local neighbors; (3) Subspace Encoding FM (SEFM)~\citep{lan2019accurate}: it captures the nonlinear relationships of individual features and feature interactions to the class label by applying explicit nonlinear feature mapping to each input feature. 

Although these FM extensions have shown improved accuracy on different predictive tasks, the space and computational complexity of these methods are much higher than the original FM. This high space and computational complexities will limit their applications in resource-constrained scenarios, such as model deployment on mobile phones or IoT devices. We seek to develop a novel memory and computational efficient FM model that can capture highly nonlinear patterns in the data. In this paper, we focus on reducing the memory cost of SEFM. The core idea of SEFM is to apply one-hot encoding for each input feature, which can effectively model the relationship of individual features and feature interactions to the class label in a highly-nonlinear way. SEFM can achieve the same computational complexity as FM for model inference. However, with respect to the space complexity of the model parameters, SEFM increases the memory cost of FM model by $b$ times where $b$ is the number of bins when applying one-hot encoding to each input feature. 

Motivated by recent works that reduce the memory cost of classification models by binarizing the model parameters~\citep{hubara2016binarized,rastegari2016xnor,alizadeh2018empirical,shen2017classification}, we propose to reduce the memory cost of SEFM by binarizing their model parameters. By constraining the model parameters of SEFM to be binary values (i.e., 1 or $-1$), only one bit is required to store each parameter value. Therefore, our proposed method can significantly reduce the memory cost of SEFM. Note that our proposed method is different from Discrete FM~\citep{liu2018discrete} (DFM) which only binarizes the model parameter for pairwise feature interactions on the original feature space. Our method based on mapped feature space will have a higher expressiveness capacity than DFM. 
In addition, DFM focuses on regression loss and their proposed training algorithm can not be directly applied for the hinge loss or logloss which are popular for classification problems. In this paper, we propose to train our binarized FM model by using Straight Through Estimator (STE) ~\citep{bengio2013estimating} with adaptive gradient descent (AdaGrad)~\citep{duchi2011adaptive} which can apply to any convex loss function.  

The contributions of this paper can be summarized as: First, we propose a novel method that can reduce the memory cost of SEFM by binarizing the model parameters. Second, since directly learning the binary model parameters is an NP-hard problem, we propose to use STE with Adagrad to effectively and efficiently obtain the binary model parameters. We provide a detailed analysis of our method's space and computational complexities for model inference and compare it with other competing methods. Finally, we evaluate the performance of our proposed method on eight datasets. Our experimental results have clearly shown that our proposed method can get much better accuracy than FM while having a similar memory cost. The results also show that our proposed method achieves comparable accuracy with SEFM but with much less memory cost.

\section{Preliminaries}\label{sec:prelim}
\textbf{Factorization Machines.} Given a training data set $D = \{\mathbf{x}_i, y_i\}_{i=1}^{n}$, where $\mathbf{x}_i$ is a $d$ dimensional feature vector representing the $i$-th training sample, $y_i \in \{-1,1\}$ is the class label of the $i$-th training sample. $x_{ij}$ denotes the $j$-th feature value of the $i$-th sample. The predicted value for a data sample $\mathbf{x}_i$ based on FM with degree-2 is,

\begin{eqnarray}\label{eq:fm}
f(\mathbf{x}_i) = \sum_{j=1}^{d}w_jx_{ij} + \sum_{j=1}^{d}\sum_{k=j+1}^{d}\tilde{w}_{jk}x_{ij}x_{ik},
\end{eqnarray}

where $w_j$ models the relationship between the $j$-th feature and the class label, and $\tilde{w}_{jk}$ models the relationship between the pairwise feature interaction of feature $j$ and feature $k$ and the class label. In FM, the model coefficient for feature interaction $\tilde{w}_{jk}$ is factorized as $\mathbf{v}_j^T\mathbf{v}_k$, where $\mathbf{v}_j$ is an $m$-dimensional vector. By using this factorization, the second term in (\ref{eq:fm}) can be computed in time $O(md)$ which is in linear with respect to the number of feature $d$ according to Lemma 3.1 in \cite{rendle2010factorization},

\begin{equation}\label{eq:fm_interaction}
\begin{split}
&\sum_{j=1}^{d}\sum_{k=j+1}^{d}\tilde{w}_{jk}x_{ij}x_{ik} = \sum_{j=1}^{d}\sum_{k=j+1}^{d}\langle \mathbf{v}_j^T\mathbf{v}_k\rangle x_{ij}x_{ik}\\
&=\frac{1}{2}\sum_{f=1}^{m}((\sum_{j=1}^{d}v_{j,f}x_{ij})^2 - \sum_{j=1}^{d}v_{j,f}^2x_{ij}^2).
\end{split}
\end{equation}

The model parameters of FMs are $\mathbf{w} \in$ $\mathbb{R}^{d}$ and $\mathbf{V} \in \mathbb{R}^{d \times m}$. These model parameters can be efficiently learnt from data by using Stochastic Gradient Descent (SGD).  
The gradient of FM prediction based on one sample $\mathbf{x}_i$ can be computed as
\begin{equation}\label{eq:fm_sgd}
  \begin{split}
    \frac{\partial f(\mathbf{x}_i)}{\partial w_j} & = x_{ij}\\
    \frac{\partial f(\mathbf{x}_i)}{\partial v_{jf}}  & = x_{ij}\sum_{j=1}^{d}v_{jf}x_{ij} - v_{jf}x_{ij}^{2}.
  \end{split}
\end{equation}

By using low rank factorization of $\tilde{w}_{jk} = \mathbf{v}_j^T\mathbf{v}_k$, FM can model all pairwise feature interations in $O(md)$ time which is much more efficient than SVM with polynomial-2 kernel \cite{chang2010training}. 

\begin{figure*}[]
\begin{center}
\subfigure[FM]{\includegraphics[width=1.53in]{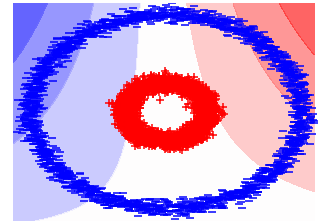}}
\subfigure[SEFM]{\includegraphics[width=1.53in]{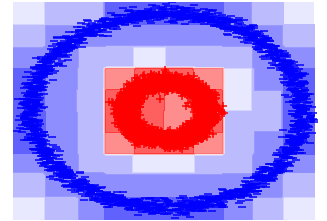}}
\subfigure[Our Method]{\includegraphics[width=1.53in]{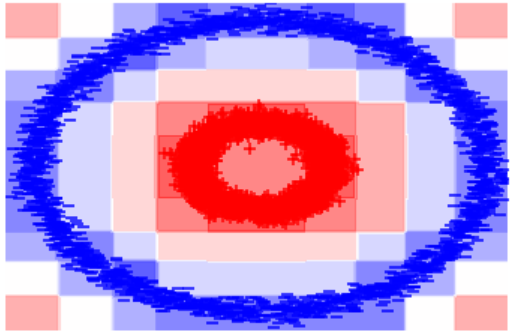}}\\
\end{center}
\caption{Comparison of the decision boundaries of FM, SEFM and our method on {\sf circles} data}
\label{fig:circles}
\end{figure*}

\noindent \textbf{Subspace Encoding Factorization Machines}. As shown in (\ref{eq:fm}), the relationship of individual features and feature interactions to the class label in FM is just linear and therefore it fails to capture highly nonlinear pattern in the data. Recently, SEFM \cite{lan2019accurate} was proposed to address the expressiveness limitation of FM. The basic idea of SEFM is to apply one-hot encoding to each individual feature. This simple idea does an efficient non-parametric nonlinear feature mapping for both individual features and feature interactions and therefore can improve the model expressiveness capacity. For a given data sample $\mathbf{x}_i$, SEFM will transform it to a very sparse vector $\mathbf{z}_i$ with length $d \times b$ by using one-hot encoding where $b$ is the number of bins that each feature is binned into. That is 

\begin{equation}\label{eq:subspace}
  z_{ijh} =
  \begin{cases}
    1,\ \text{if}\ x_{ij} \in B_h^j\\
    0,\ \text{otherwise},
  \end{cases}
\end{equation}

where $B_h^j$ denotes the interval boundary of the $h$-bin of the $j$-th feature in original space. In other words, original feature matrix $\mathbf{X} \in \mathbb{R}^{n \times d}$ will be transformed to a binary feature matrix $\mathbf{Z} \in \mathbb{R}^{n \times p}$ where $p = d \times b$. Then a FM model is trained on the new data $\{\mathbf{z}_i, y_i\}_{i=1}^{n}$. The predicted value of SEFM will be $\mathbf{w} \in$ $\mathbb{R}^{p}$ and $\mathbf{V} \in \mathbb{R}^{p \times m}$ and $p$ is $b$ times larger than $d$.

\begin{eqnarray}\label{eq:fm_z}
f(\mathbf{z}_i) = \sum_{j=1}^{p}w_jz_{ij} + \sum_{j=1}^{p}\sum_{k=j+1}^{p}(\mathbf{v}_j^T\mathbf{v}_k)z_{ij}z_{ik}.
\end{eqnarray}

Figure \ref{fig:circles} demonstrates the comparison of the decision boundaries between FM and SEFM on {\sf circles} data, which is a nonlinear synthetic classification dataset in two-dimensional space. The blue points in the large outer circle belong to one class and the red points in the inner circle belong to the other class. As shown in sub-figure (a), the nonlinear decision boundary based on polynomial feature expansion produced by FM cannot capture the highly nonlinear pattern. SEFM (sub-figure (b)) produces the piecewise axis perpendicular decision boundary which works well on this data. The sub-figure (c) shows the decision boundary of our proposed method which also works very well on this nonlinear data but with much less memory cost compared with SEFM. 

\section{Methodology}
\subsection{Binarized Factorization Machines}
Even though SEFM improves the model expressiveness capacity of FM and therefore achieves higher accuracy than FM on highly nonlinear datasets, SEFM increases the memory cost of FM by $b$ times since the model parameters are $\mathbf{w} \in$ $\mathbb{R}^{p}$ and $\mathbf{V} \in \mathbb{R}^{p \times m}$ where $p$ is $b$ times larger than the original data dimensionality $d$. The high space complexity of SEFM will limit its applications in resource-constrained scenarios, such as model deployment on mobile phones or IoT devices. 

To reduce the memory cost of SEFM, we propose to binarize the model parameter $\mathbf{w}$ and $\mathbf{V}$. We approximate the full-precision $\mathbf{w}$ and $\mathbf{V}$ using binarized $\mathbf{w}^b \in \{-1, 1\}^{p}$ and $\mathbf{V}^b \{-1,1\}^{p \times m} $ and scaling parameters $\alpha$, $\beta$. That is, $\mathbf{w} \approx \alpha\mathbf{w}^b$ and $\mathbf{V} \approx \beta\mathbf{V}^b$. The predicted value of our model will be
\begin{equation}
f(\mathbf{z}_i) = \alpha\sum_{j=1}^{p}w^b_jz_{ij} + \beta^2\sum_{j=1}^{p}\sum_{k=j+1}^{p}((\mathbf{v}^b_j)^T\mathbf{v}^b_k)z_{ij}z_{ik}. 
\end{equation}
In our experiment, we show that using the scaling parameter $\alpha$ and $\beta$ can get better accuracy than without using them. The objective of our proposed model can be formulated as
\begin{equation}\label{eq:bfm}
	\begin{split}
	\min\limits_{\alpha, \beta, \mathbf{w}^b,\mathbf{V}^b}& \frac{1}{n}\sum_{i=1}^{n}\ell(y_i, f(\mathbf{z}_i)) + \frac{\lambda_1}{2}\|\alpha\mathbf{w}^b\|_2^2 + \frac{\lambda_2}{2}\|\beta\mathbf{V}^b\|_F^2 \\
	\text{s.t.} & \ \ \ \ \mathbf{w}^b \in \{-1, 1\}^{p} , \\
                & \ \ \ \ \mathbf{V}^b \in \{-1,1\}^{p \times m}, \\
                &  \ \ \ \ \alpha, \beta > 0,
	\end{split}
\end{equation}
\noindent where $\ell(y_i, f(\mathbf{z}_i))$ in (\ref{eq:bfm}) denotes a convex loss function. Due to the non-smooth binary constraints, optimization of (\ref{eq:bfm}) is an NP-hard problem and it will need $O(2^{(p+mp)})$ time to get the global optimal solution. Next, we will show how to optimize (\ref{eq:bfm}) efficiently using STE with Adagrad.

\subsection{Learning Binary Model Parameters by Straight Through Estimator (STE)}
In this section, we describe our proposed algorithm to solve (\ref{eq:bfm}) based on Straight Through Estimator (STE)~\citep{bengio2013estimating}. STE is popularly used for training binary convolution neural networks~\citep{alizadeh2018empirical}. The idea of STE is to learn a full-precision variable as the proxy of the binary variable during the training process. As in our proposed model, we introduce a full-precision $\mathbf{\widetilde{w}}$ as the proxy of binary $\mathbf{w}^b$ in (\ref{eq:bfm}) and a full-precision $\mathbf{\widetilde{V}}$ as the proxy of binary $\mathbf{V}^b$ in (\ref{eq:bfm}). Instead of directly learning binary variables $\mathbf{w}^b$ and $\mathbf{V}^b$, we learn their full-precision proxy variables $\mathbf{\widetilde{w}}$ and $\mathbf{\widetilde{V}}$. To estimate the binary variables $\mathbf{w}^b$ and $\mathbf{V}^b$ and scaling parameter $\alpha$, $\beta$, we propose to minimize $\|\mathbf{\widetilde{w}} - \alpha\mathbf{w}^b\|^2$ and $\|\mathbf{\widetilde{V}} - \beta\mathbf{V}^b\|^2$. 



Here we present our detailed derivation of estimating $\mathbf{w}^b$, $\mathbf{V}^b$ and scaling parameter $\alpha$, $\beta$ based on the full-precision proxy variables $\mathbf{\widetilde{w}}$ and $\mathbf{\widetilde{V}}$. The derivation follows the similar procedures of optimizing binary convolution neural networks \citep{rastegari2016xnor}. In the STE framework, we use SGD to update full-precision proxy variables $\mathbf{\widetilde{w}}$ and $\mathbf{\widetilde{V}}$ instead of $\mathbf{w}^b$, $\mathbf{V}^b$. We propose to get an optimal approximation for $\mathbf{w} \approx \alpha\mathbf{w}^b$ and $\mathbf{V} \approx \beta\mathbf{V}^b$ by minimizing $\|\mathbf{\widetilde{w}} - \alpha\mathbf{w}^b\|^2$ and $\|\mathbf{\widetilde{V}} - \beta\mathbf{V}^b\|^2$.

Let us consider minimize $\|\mathbf{\widetilde{w}} - \alpha\mathbf{w}^b\|^2$ first. By expanding the equation$\|\mathbf{\widetilde{w}} - \alpha\mathbf{w}^b\|^2$, it can be rewritten as 

\begin{equation}\label{eq:min_w}
\|\mathbf{\widetilde{w}} - \alpha\mathbf{w}^b\|^2  = \mathbf{\widetilde{w}}^T\mathbf{\widetilde{w}} - 2\alpha \mathbf{{\widetilde{w}}}^T\mathbf{w}^b + \alpha^2 (\mathbf{w}^b)^T\mathbf{w}^b.
\end{equation}
Since $\mathbf{w}^b \in \{-1, 1\}^{p}$, $(\mathbf{w}^b)^T\mathbf{w}^b$ will equals to a constant $p$. 
The scaling parameter $\alpha > 0$ and $\widetilde{w}$ is a known variable in (\ref{eq:min_w}). Therefore, to minimize (\ref{eq:min_w}) is equivalent to the following optimization problem,
\begin{equation}\label{eq:max_w}
\begin{split}
\max\limits_{\mathbf{w}^b}\ \ & \mathbf{{\widetilde{w}}}^T\mathbf{w}^b \\
\text{s.t.} & \ \ \ \ \mathbf{w}^b \in \{-1, 1\}^{p}. \\
\end{split}
\end{equation}
It is straightforward to the optimal solution for (\ref{eq:max_w}), which is $\mathbf{w}^b = \text{sign}({\mathbf{\widetilde{w}}})$. To obtain the optimal solution for the scaling parameter $\alpha$, we take the derivative of (\ref{eq:min_w}) with respective to $\alpha$. It is $ -2 \mathbf{{\widetilde{w}}}^T\mathbf{w}^b + 2\alpha p$. By setting it to 0, we will get $\alpha = \frac{\mathbf{{\widetilde{w}}}^T\mathbf{w}^b}{p}$. Since $\mathbf{w}^b = \text{sign}({\mathbf{\widetilde{w}}})$, the scaling parameter $\alpha$ will be

\begin{equation}\label{eq:alpha}
\alpha = \frac{\sum_j |\widetilde{w}_j|}{p}.
\end{equation}

By following the same procedure, we can also get the optimal $\mathbf{V}^b = \text{sign}({\mathbf{\widetilde{V}}})$ and the optimal $\beta$ as

\begin{equation}\label{eq:beta}
\beta = \frac{\sum_{i,j}|\widetilde{v}_{ij}| }{p\times m}.
\end{equation}

The optimal binary $\mathbf{w}^b$ is obtained by $\mathbf{w}^b = \text{sign}({\mathbf{\widetilde{w}}})$, where sign() is the element-wise sign function which return 1 if the element is larger or equal than zero and return $-1$ otherwise. Similarly, $\mathbf{V}^b$ can be obtained by $\mathbf{V}^b = \text{sign}({\mathbf{\widetilde{V}}})$.
Since the sign() function 
are not differentiable, STE estimates the gradients with respect to $\mathbf{\widetilde{w}}$ and $\mathbf{\widetilde{V}}$ as if the non-differentiable function sign() is not present. In other word, STE will simply estimate $\frac{\partial \text{sign}(\tilde{w}_j)}{\partial \tilde{w}_j} $ as $\frac{\partial \tilde{w}_j}{\partial \tilde{w}_j} = 1$. In practice, we also employ the gradient clipping as in~\citep{hubara2016binarized}. Then, the gradient for the non-differentiable sign function is    
\begin{equation}\label{eq:gradientClip}
   \frac{\partial \text{sign}(\tilde{w}_j)}{\partial \tilde{w}_j} = 1_{|\tilde{w}_j|\le 1}. 
\end{equation}

Therefore, the gradient of (\ref{eq:bfm}) with respect to $\tilde{w}_j$ (i.e., the $j$-th element in the proxy variable $\mathbf{\tilde{w}}$) based on the $i$-th data sample can be estimated as 
\begin{equation}\label{eq:proxy_gradient_w_j}
\begin{split}\begin{aligned}
g_{\tilde{w}_j}&= \frac{\partial \ell(y_i, f(\mathbf{z}_i))}{\partial \tilde{w}_j} = (\frac{\partial \ell(y_i, f(\mathbf{z}_i)) }{\partial w^b_j} + \lambda_1\alpha w^b_j)\frac{\partial w^b_j}{\partial \tilde{w}_j} \\
 & = (\frac{\partial \ell(y_i, f(\mathbf{z}_i)) }{\partial f(\mathbf{z}_i)}\alpha z_{ij} + \lambda_1\alpha w^b_j) 1_{|\tilde{w}_j|\le 1}.
\end{aligned}\end{split}
\end{equation}
Any convex loss function can be used in (\ref{eq:proxy_gradient_w_j}) and the gradient $\frac{\partial \ell(y_i, f(\mathbf{z}_i))}{\partial f(\mathbf{z}_i)}$ can be easily obtained for different loss function. 

Similarly, the gradient of (\ref{eq:bfm}) with respect to $\tilde{v}_{jf}$ (i.e., the $(j,f)$-th element in $\mathbf{\widetilde{V}}$) based on the $i$-th data sample can be estimated as 

\begin{equation}\label{eq:proxy_gradient_V_jf}
g_{\tilde{v}_{jf}} = (\frac{\partial \ell(y_i, f(\mathbf{z}_i)) }{\partial f(\mathbf{z}_i)}\frac{\partial f(\mathbf{z}_i)}{v^b_{jf}} + \lambda_2\beta v^b_{jf}) 1_{|\tilde{v}_{jf}|\le 1}.
\end{equation}
where $\frac{\partial f(\mathbf{z}_i)}{v^b_{jf}}$ 
can be computed based a similar form as in (\ref{eq:fm_sgd}). That is
\begin{equation}\label{eq:gradient_vij}
    \frac{\partial f(\mathbf{z}_i)}{\partial v^b_{jf}} = \beta^2(z_{ij}\sum_{j=1}^{p}v^b_{jf}z_{ij} - v^b_{jf}z_{ij}^{2}).
\end{equation}

\noindent Note that ($\ref{eq:gradient_vij}$) can be computed in $O(1)$ time if precomputing $\sum_{f=1}^{p}v^b_{jf}z_{ij}$. Based on the stochastic gradient as shown in (\ref{eq:proxy_gradient_w_j}) and (\ref{eq:proxy_gradient_V_jf}), we can update $\mathbf{\widetilde{w}}$ and $\mathbf{\widetilde{V}}$ using SGD method. 

\subsection{Updating the Proxy Variables by Adagrad}
\begin{figure}[H]
\begin{center}
\begin{tabular}{c}
\subfigure[banana]{\includegraphics[width=2.2in]{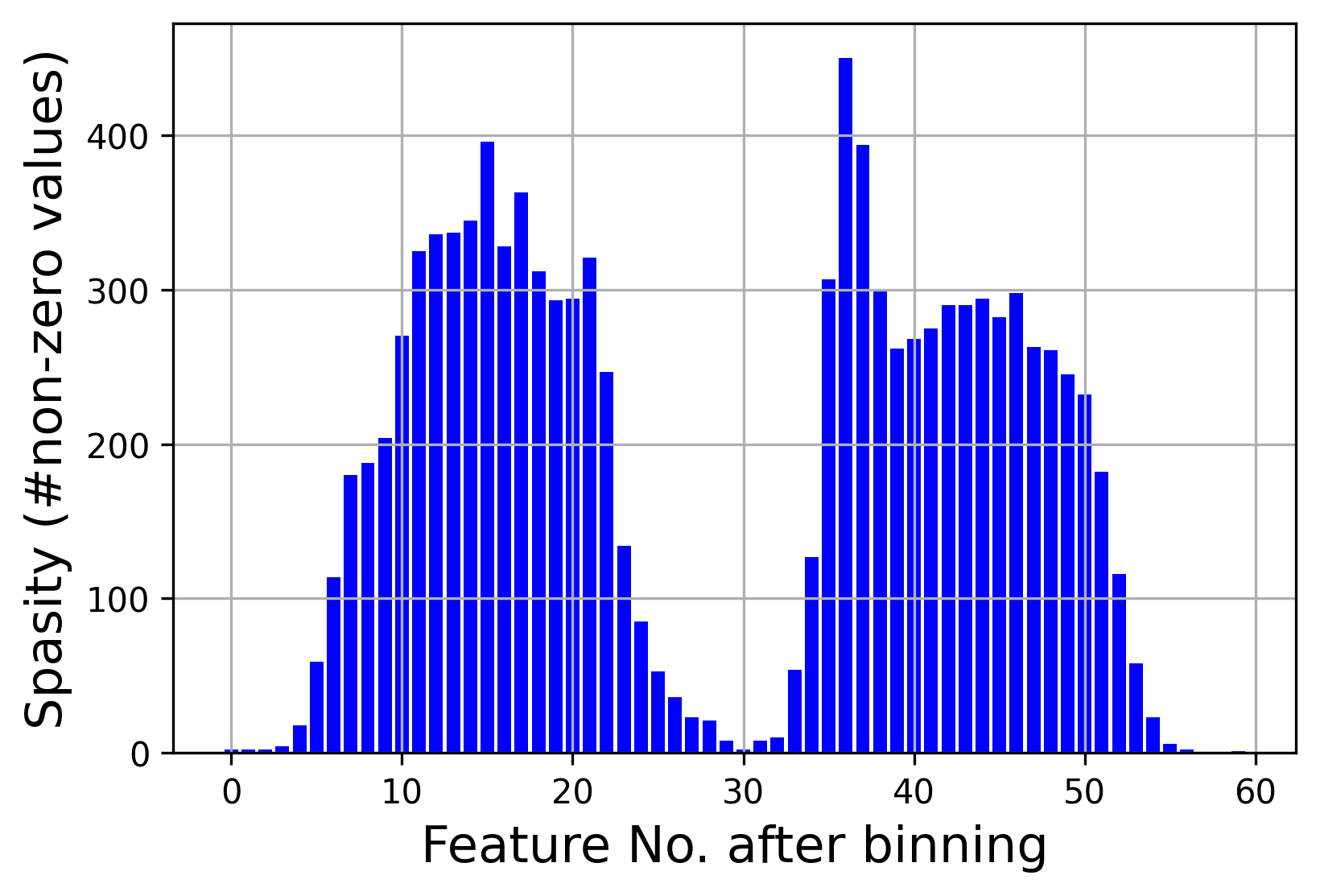}}\\
\subfigure[pendigits]{\includegraphics[width=2.2in]{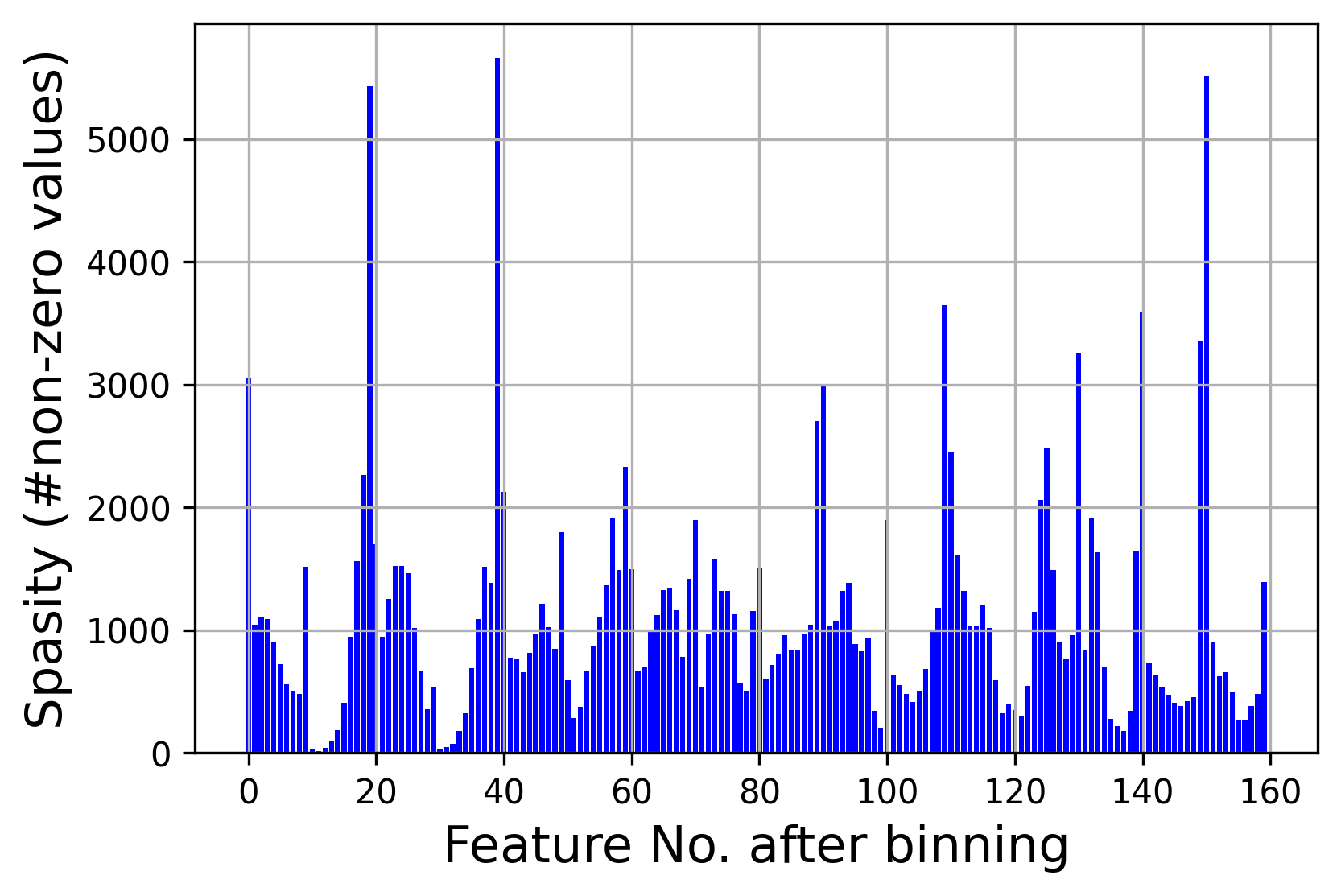}}\\
\subfigure[ijcnn]{\includegraphics[width=2.2in]{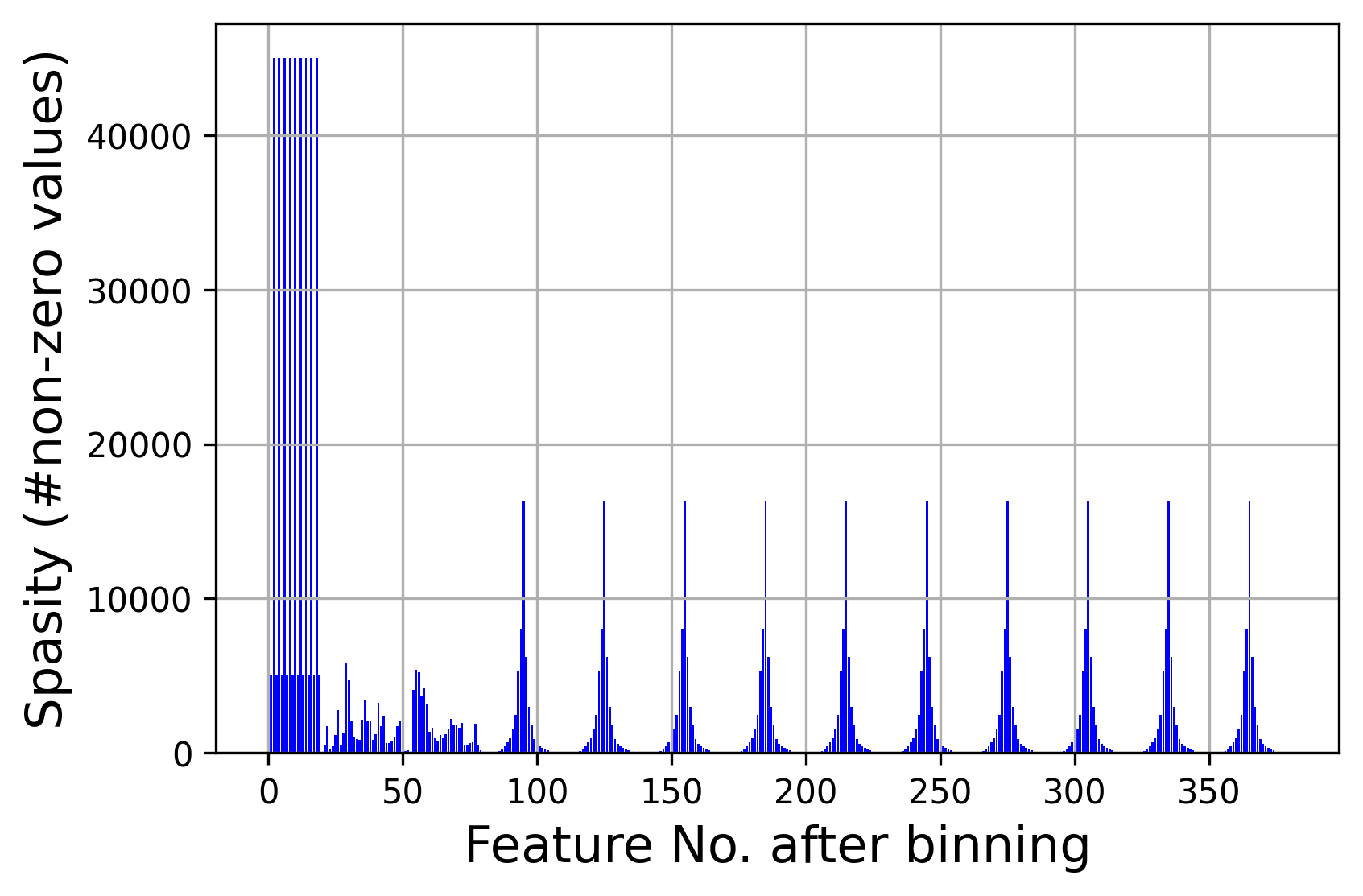}}
\end{tabular}
\end{center}
\caption{Feature sparsity after binning on different datasets}
\label{fig:feature_sparsity}
\end{figure}

In our objective optimization problem (\ref{eq:bfm}), the feature representation $\mathbf{Z}$ after binning is very sparse since it is generated by applying one-hot encoding to each original feature. Also, the sparsity rate for different features in $\mathbf{Z}$ will be very different, as shown in Figure \ref{fig:feature_sparsity}. Therefore, parameters associated with features with high sparsity rate will be updated less frequently than parameters associated with features with low sparsity rate in SGD. If we directly applied the standard SGD where the same learning rate $\eta$ is set for all parameters, it could be either too small for the parameters associated with features with high sparsity rate or too large for the parameters associated with features with low sparsity rate. 

To address this issue, we propose to use Adagrad~\citep{duchi2011adaptive} to update proxy variables $\mathbf{\widetilde{w}}$ and $\mathbf{\widetilde{V}}$. Adagrad uses different learning rates for different model parameters and the learning rates can be adaptively adjusted during the training process. It has been shown that Adagrad can greatly improve the robustness of SGD~\citep{ruder2016overview}.
The basic intuition of Adagrad is to give low learning rates for parameters associated with features with low sparsity rate and to given high learning rates for parameters associated with features with high sparsity rate. 

For a parameter (e.g., $\tilde{w}_j$ or $\tilde{v}_{jf}$) in our model, based on the idea of Adagrad, we set the learning rate in the $t$-th update $\eta_t$ to $\frac{\eta}{\sqrt{s_t} + \epsilon}$ where $s_t$ equals the sum of squares of previously observed gradients up to time $t$ and $\epsilon$ is a small constant that prevents division by 0. 

Therefore, for updating the $j$-th element in $\mathbf{\tilde{w}}$, we can first obtain the gradient $g_{\tilde{w}_j}$ as in (\ref{eq:proxy_gradient_w_j}). Then, the updating rule for $\tilde{w}_j$ at the $t$-th time will be 

\begin{equation}\label{eq:adagrad_w}
\tilde{w}_j^{t+1}  = \tilde{w}_j^{t} - \frac{\eta}{\sqrt{s_j^{t} + \epsilon}}g_{\tilde{w}_j}, 
\end{equation}
where $s_j^t$ aggregates sum of squares of previously observed gradients up to time $t$. That is, $s_j^{t}  = s_j^{t-1} + (g_{\tilde{w}_j})^2.$

Similarly, for updating the $(j,f)$-th parameter in $\mathbf{\tilde{V}}$, we first obtain the gradient $g_{\tilde{v}_{jf}}$ as in (\ref{eq:proxy_gradient_V_jf}). Then, the updating rule for $\tilde{v}_{jf}$ at the $t$-th time will be     
\begin{equation}\label{eq:adagrad_V}
    \tilde{v}_{jf}^{t+1}  = \tilde{v}_{jf}^{t} - \frac{\eta}{\sqrt{\tilde{s}_{jf}^{t} + \epsilon}}g_{\tilde{v}_{jf}}, 
\end{equation}
where $\tilde{s}_{jf}^t$ aggregates sum of squares of previously observed gradients with respect to $\tilde{v}_{jf}$ up to time $t$. That is, $\tilde{s}_{jf}^{t}  = \tilde{s}_{jf}^{t-1} + (g_{\tilde{v}_{jf}})^2.$

As shown in (\ref{eq:adagrad_w}) and (\ref{eq:adagrad_V}), Adagrad adaptively adjusts the learning rate $\eta$ at each time step $t$ for each parameter in $\mathbf{\tilde{w}}$ and $\mathbf{\tilde{V}}$ based on their past gradients. In our experiment, we have shown that Adagrad can achieve lower training loss and is more robust than SGD.

\begin{algorithm}[tb]
  \caption{Learning Binary Model Parameters by Straight Through Estimator (STE) with Adagrad}
  \begin{algorithmic}\normalsize
  \State \underline{\textbf{Training}}
  \State \textbf{Input}: training data set $D = \{\mathbf{x}_i, y_i\}_{i=1}^{n}$, low-rank parameter $m$, number of bins $b$, step size $\eta$, regularization parameters $\lambda_1$, $\lambda_2$ and a small constant $\epsilon$;
  \State \textbf{Output}: Binarized FM model {$\mathbf{w}^b \in \{-1, 1\}^{p},\mathbf{V}^b \in \{-1, 1\}^{p \times m}$} on mapped feature space;
  \end{algorithmic}
  \begin{algorithmic}[1]\normalsize
  \State Generate new feature representation $\mathbf{Z}$ 
  \State Initialize proxy variables  
  $\mathbf{\tilde{w}} = \mathbf{0}$, $\mathbf{\tilde{V}} = \mathbf{0}$ and $\mathbf{s} \in \mathbb{R}^{p} = \mathbf{0}$, $\tilde{\mathbf{S}} \in \mathbb{R}^{p \times m} = \mathbf{0}$  which are used for adjusting learning rate in Adagrad. Use (\ref{eq:alpha}) and (\ref{eq:beta}) to set scaling parameter $\alpha$, $\beta$.
  \Repeat 
  \For{$\forall i \in [1,n]$}
              \For{\text{each nonzero $z_{ij}$}}
                 \State compute the gradient $g_{\tilde{w}_j}$ as in (\ref{eq:proxy_gradient_w_j})
                 \State update $s_j$ as $s_j\mathrel{+}=(g_{\tilde{w}_j})^2$
                 \State update $\tilde{w}_j$ as ${\tilde{w}}_j  = {\tilde{w}}_j - \frac{\eta}{\sqrt{s_j + \epsilon}}g_{\tilde{w}_j}$
                 \State ${w}^b_j = \text{sign}({\tilde{w}_j})$
             \EndFor
             \For{$\forall f \in [1,m]$}
                 \For{\text{each nonzero $z_{ij}$}}
                     \State compute the gradient $g_{\tilde{v}_{jf}}$ as in (\ref{eq:proxy_gradient_V_jf})
                     \State update $\tilde{s}_{jf}$ as $\tilde{s}_{jf}\mathrel{+}=(g_{\tilde{v}_{jf}})^2$
                     \State $\tilde{v}_{jf}  = {\tilde{v}}_{jf} - \frac{\eta}{\sqrt{\tilde{s}_{jf} + \epsilon}} g_{\tilde{v}_{jf}}$
                     \State ${v^b_{jf}} = \text{sign}(\tilde{v}_{jf})$
                 \EndFor
             \EndFor
  \EndFor
  \State update scaling parameter $\alpha$, $\beta$ using (\ref{eq:alpha}) and (\ref{eq:beta})
  \Until{\emph{stopping criterion is met}}
  \end{algorithmic}
  \label{alg:on_deviceFM}
\end{algorithm}

\begin{table}[h]
  \caption{Memory costs for different FM models}
      \begin{minipage}[b]{1\linewidth}\centering \normalsize
            \begin{tabular}{lr}
            \toprule
            {\sf Method} & {\sf Memory Cost (bits)} \\
            \midrule
            FM & 32 $\times$ ( $d$ + $m$ $\times$ $d$ ) \\
            SEFM & 32 $\times$ ( $d$ $\times$ $b$ + $m$ $\times$ $d$ $\times$ $b$ ) \\
            DFM & 32 $\times$ $d$ + $m$ $\times$ $d$  \\
            Binarized FM & $d$ $\times$ $b$ + $m$ $\times$ $d$ $\times$ $b$  \\
            \bottomrule
            \end{tabular}
        \end{minipage}
        
        \label{table:memory_costs}
\end{table}

\begin{table}[ht!]
    \centering
    \caption{The accuracy, prediction time and memory cost of different methods
        }
        
    \rotatebox{90}{
            \begin{tabular}{llcccccc}
            \toprule
            Dataset & Performance & { Liblinear} & { FM} & { DFM} & { NFM} &{ SEFM}  & {Binarized FM} \\
             $n$/$d$/\#class & &         &        &         &               &          &  \\
           \midrule
            {\sf circles} & accuracy (\%) & 47.88$\pm$1.75 & 49.20$\pm$1.29 & 51.03$\pm$2.94 & \textbf{99.95$\pm$0.04} & 99.91$\pm$0.08  & \textbf{99.95$\pm$0.06} \\
            5,000/2/2 & prediction time (ms)  & 0.3 & 2.0 & 2.2 & 39.3 & 1.6    & 1.6 \\
             & memory cost & 0.06$\times$ & 1$\times$ & 0.12$\times$ & 893$\times$ & 30$\times$  & 0.94$\times$ \\
            \hline
            {\sf moons} & accuracy (\%) & 89.68$\pm$0.42 & 88.57$\pm$0.37 & 88.71$\pm$0.96 & 99.93$\pm$0.07 & \textbf{99.99$\pm$0.00}  & \textbf{99.99$\pm$0.00}  \\
            5,000/2/2 & prediction time (ms) & 0.3 & 2.7 & 2.8 & 92.7 &  1.8   & 1.2  \\
             & memory cost & 0.01$\times$ & 1$\times$ & 0.02$\times$ & 130$\times$ & 3.95$\times$  & 0.12$\times$ \\
            \hline
            {\sf banana} & accuracy (\%) & 56.10$\pm$0.58 & 54.82$\pm$3.65 & 56.74$\pm$1.12 & 68.49$\pm$1.88 & \emph{88.98$\pm$0.47}  & \textbf{89.25$\pm$0.48}  \\
             5,300/2/2 & prediction time (ms) & 0.3 & 1.7 & 2.0 & 132.2 & 1.3  & 1.8   \\
             & memory cost & 0.06$\times$ & 1$\times$ & 0.09$\times$ & 1849$\times$ & 30$\times$  & 3.66$\times$  \\
            \hline
            \hline
            {\sf breast-cancer} & accuracy (\%) & 95.02$\pm$0.94 & 94.73$\pm$0.64 & 95.22$\pm$1.26 & \textbf{96.91$\pm$0.74} & 95.51$\pm$0.94  & \emph{96.20$\pm$0.64}  \\
             683/10/2 & prediction time (ms)  & 0.2 & 0.9  & 0.8 & 13.0 & 0.6  & 0.6  \\
             & memory cost & 0.06$\times$ & 1$\times$ & 0.12$\times$ & 300$\times$ & 30$\times$  & 3.58$\times$ \\
            \hline
            {\sf segment} & accuracy (\%) & 91.52$\pm$1.03 & 91.98$\pm$1.60 & 90.51$\pm$1.09 & 88.17$\pm$0.87 & \emph{94.23$\pm$0.82}  & \textbf{94.75$\pm$0.82}  \\
            2,310/19/7 & prediction time (ms) & 0.3 & 2.6 & 4.1 & 136.5 & 30.1  & 16.4 \\
             & memory cost & 0.06$\times$ & 1$\times$ & 0.18$\times$ & 484$\times$ & 19.4$\times$  & 2.37$\times$  \\
            \hline
            {\sf pendigits} & accuracy (\%) & 93.12$\pm$0.36 & 94.26$\pm$0.61 & 95.74$\pm$0.45 & 94.29$\pm$0.15 & \emph{97.38$\pm$0.12}   & \textbf{97.56$\pm$0.16} \\   
           10,992/16/10 & prediction time (ms) & 0.9 & 168.2  & 36.6 & 2,045.1 & 1,268.8    & 1,700.3   \\
             & memory cost & 0.01$\times$ & 1$\times$ & 0.04$\times$ & 53$\times$ & 10$\times$  & 0.31$\times$   \\
            \hline
            {\sf ijcnn}   & accuracy (\%) & 92.08$\pm$0.20 & 94.75$\pm$0.26 & 94.59$\pm$0.35 & 95.61$\pm$0.08 & \emph{96.29$\pm$0.18}  & \textbf{96.45$\pm$0.09} \\
             49,990/22/2 & prediction time (ms)  & 8.9 & 158.3  & 193.7 & 651.2 &  125.7   & 229.2 \\
             & memory cost & 0.01$\times$ & 1$\times$ & 0.04$\times$ & 45$\times$ & 7.67$\times$  & 0.94$\times$ \\
            \hline
            {\sf webspam} & accuracy (\%) & 92.71$\pm$0.04 & 95.49$\pm$0.72 & 95.21$\pm$0.15 & 97.53$\pm$0.12 & \textbf{98.11$\pm$0.13}  & \emph{97.79$\pm$0.08}  \\
            350,000/254/2 & prediction time (ms) & 67.4 & 5,025.2   & 4,997.9 & 54,605.7 & 7,792.2   & 7,004.4   \\
             & memory cost & 0.01$\times$ & 1$\times$ & 0.04$\times$ & 2.06$\times$ & 30$\times$  & 0.94$\times$  \\
       \bottomrule
            \end{tabular}
    }
        
    \label{table:evaluation_results}

\end{table}

\subsection{Algorithm Implementation and Analysis}
We summarize our proposed method in Algorithm \ref{alg:on_deviceFM}. The step 1 obtains the new feature representation $\mathbf{Z}$ by applying one-hot encoding on each feature in input data $\mathbf{X}$ which can be done in $O(nd)$ time~\citep{lan2019accurate}. The step 2 initializes the proxy variables $\mathbf{\tilde{w}}$, $\mathbf{\tilde{V}}$ and $\mathbf{s} \in \mathbb{R}^{p} = \mathbf{0}$, $\tilde{\mathbf{S}} \in \mathbb{R}^{p \times m} = \mathbf{0}$ which are used for adjusting learning rate in Adagrad. From step 3 to 21, we update the proxy parameters and the binary parameters using STE with Adagrad. The time complexity for learning the binarized parameter is $O(nnz(\mathbf{Z})ml)$, where 
$nnz(\mathbf{Z})$ denotes the number of non-zero values in $\mathbf{Z}$, $m$ is the low-rank parameter and $l$ is the number of passes over dataset $\mathbf{Z}$. Empirically, the algorithm converge very fast and $l$ is a small number in our experiments.


\subsection{Model Inference of Our Proposed Model}
Similar to Lemma 3.1 in~\citep{rendle2010factorization}, 
the prediction score of our model for a given input $\mathbf{z}_i$ can be written as 

\begin{equation}\label{eq:bfm_prediction}
\begin{split}\begin{aligned}
& f(\mathbf{z}_i)  = \alpha\sum_{j=1}^{p}w^b_jz_{ij} + \beta^2\sum_{j=1}^{p}\sum_{k=j+1}^{p}\langle (\mathbf{v}^b_j)^T\mathbf{v}^b_k\rangle z_{ij}z_{ik}\\
& = \alpha\sum_{j=1}^{p}w^b_jz_{ij} + \frac{1}{2}\beta^2\sum_{f=1}^{m}((\sum_{j=1}^{p}v^b_{jf}z_{ij})^2 - \sum_{j=1}^{p} {v^b_{jf}}^2 z_{ij}^2)\\
\end{aligned}\end{split}
\end{equation}

Since each element in $\mathbf{V}^b$ is either 1 or $-1$, ${v^b_{jf}}^2 = 1$ and $\sum_{j=1}^{p} {v^b_{jf}}^2 z_{ij}^2 = nnz(\mathbf{z}_i)$ where $nnz(\mathbf{z}_i)$ is the number of non-zero values in $\mathbf{z}_i$ which will be equal to $d$ because of one-hot encoding. Also note that $\mathbf{w}^b$ and $\mathbf{V}^b$ are binary weights and $\mathbf{z}_i$ is a sparse vector with $d$ nonzero values. The nonzero value in $z_i$ can only be 1. Therefore, the model inference as shown in (\ref{eq:bfm_prediction}) can be estimated by very efficiently only using addition and subtraction operations (without the need of multiplication operation). Also the arithmetic operations in (\ref{eq:bfm_prediction}) can be further replaced cheap XNOR and POPCOUNT bitwise operations~\citep{rastegari2016xnor}.  

During model inference, we do not need to keep the proxy variables $\mathbf{\tilde{w}}$ and  $\mathbf{\tilde{V}}$. We only need the binary parameters $\mathbf{w}^b$, $\mathbf{V}^b$ and scaling parameter $\alpha$, $\beta$ to compute the prediction score for a test sample based on (\ref{eq:bfm_prediction}). We summarize the memory cost for different FM models in Table \ref{table:memory_costs}.

\section{Experiments}

In this section, we compare our proposed method with other competing algorithms on eight datasets. The summary information of each dataset is given in the first column of Table \ref{table:evaluation_results}. Among them, there are three synthetic datasets and five real-life datasets. These three synthetic datasets (i.e., {\sf circles}, {\sf moons} and {\sf banana}) are widely used for evaluation nonlinear models. For the real-life datasets, three of them are binary classification and two of them are multi-class classification. One-vs-all strategy is used for multi-class classification. 

In our experiments, we evaluate the performance of the following six different classification methods:

\begin{itemize}[leftmargin=*]
\item {\bf Liblinear:} an efficient solver for linear support vector machine~\citep{fan2008liblinear}.
\item {\bf FM:} The original Factorization Machines~\citep{rendle2010factorization}. We use the implementation of fastFM~\citep{bayer2016fastfm}.
\item {\bf DFM:} Discrete Factorization Machines~\citep{liu2018discrete} which binarize the feature interaction parameters in FM. The DFM model focused on regression loss, and we extend it for classification loss. 
\item {\bf NFM:} Neural Factorization Machines~\citep{he2017neural} which combines the FM with the neural networks.
\item {\bf SEFM:} It improves expressive limitation of FM by one-hot encoding to each input features~\citep{lan2019accurate}.  
\item {\bf Binarized FM:} our proposed method. 
\end{itemize}

\subsection{Experimental Setting.}
For each dataset, we randomly select 70\% as training data and use the remaining 30\% as test data. This procedure is repeated for 10 times and we report the average accuracy on test data. The low-rank parameter $m$ for FM, SEFM, DFM and Binarized FM is chosen from \{16, 32, 64, 128\}. The parameter $b$ (i.e., the number of bins) of SEFM and Binarized FM is chosen from \{10, 20, 30, 40, 50\}. The regularization parameter in Liblinear, FM, DFM and SEFM is chosen from $\{10^{-2}, 10^{-1}, \dots, 10^{2}\}$. The optimal parameter combination is selected by 5-fold cross-validation on training data. The suggested settings are used in NFM, but for {\sf webspam}, the hidden factor and the number of layers are both decreased to make it successfully run in our computing environment.

\begin{figure*}[]
\begin{center}
\subfigure[circles]{\includegraphics[width=2.2in]{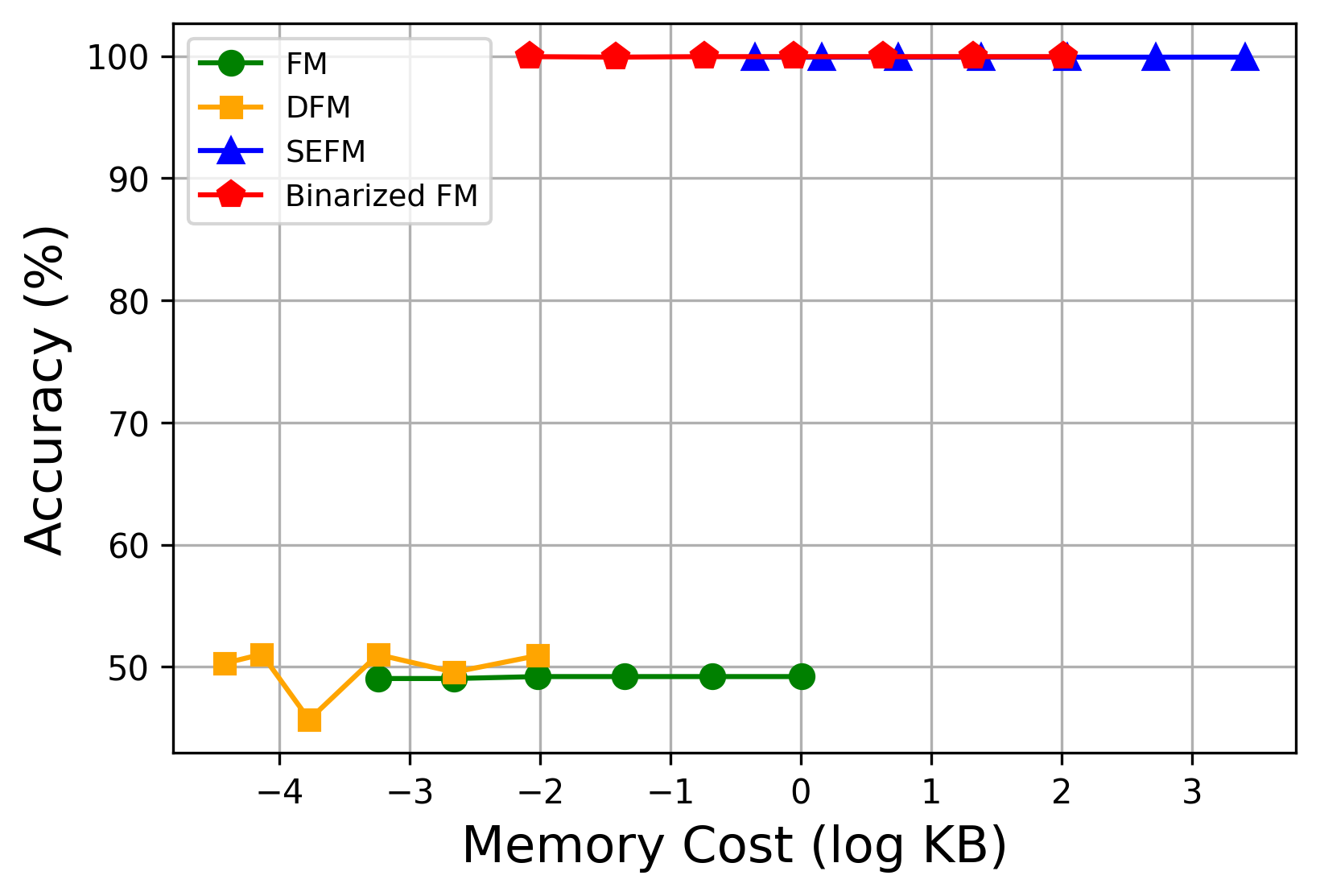}}
\hspace{0.05in}
\subfigure[moons]{\includegraphics[width=2.2in]{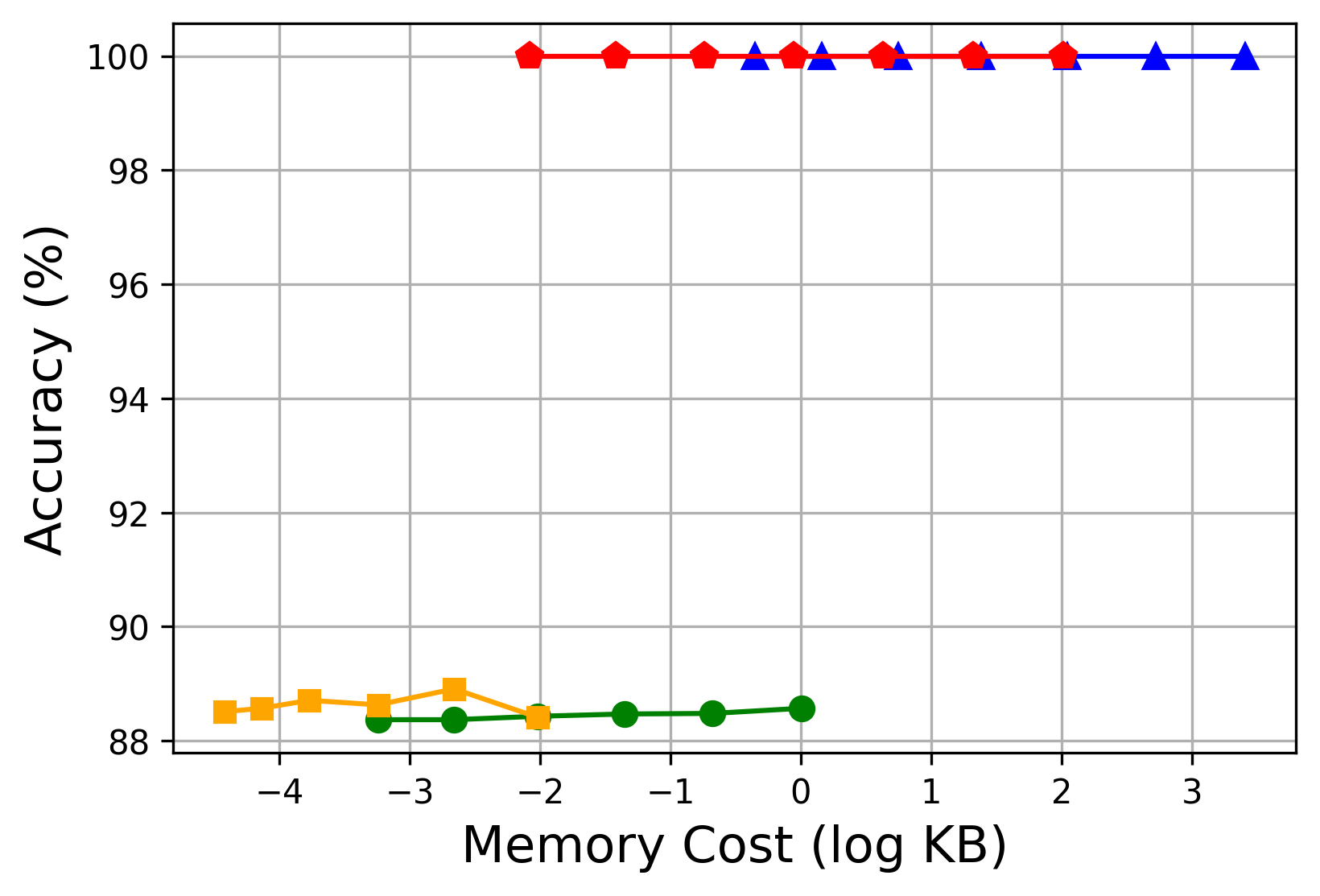}}
\\
\vspace{0.1in}
\subfigure[segment]{\includegraphics[width=2.2in]{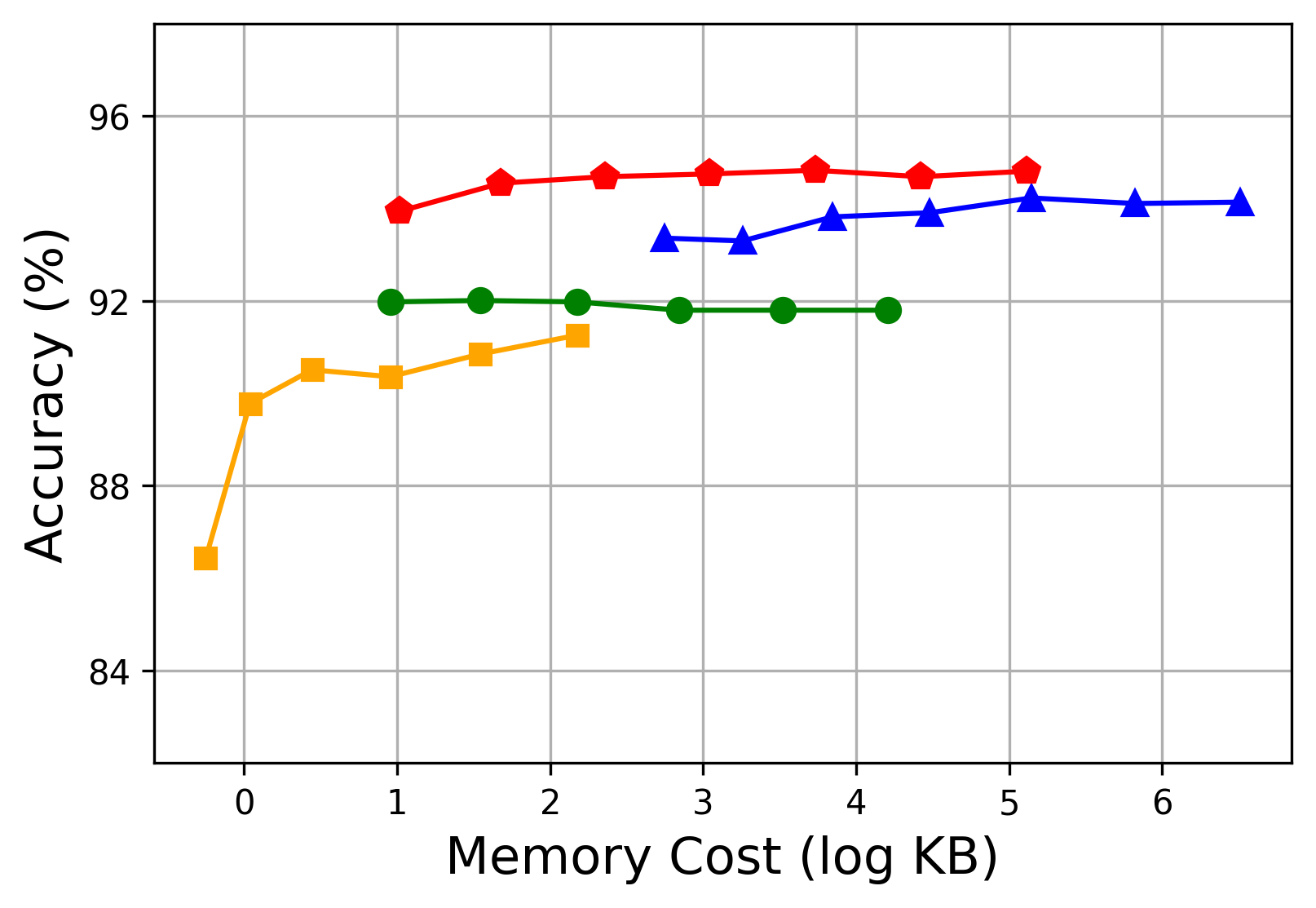}}
\hspace{0.05in}
\subfigure[pendigits]{\includegraphics[width=2.2in]{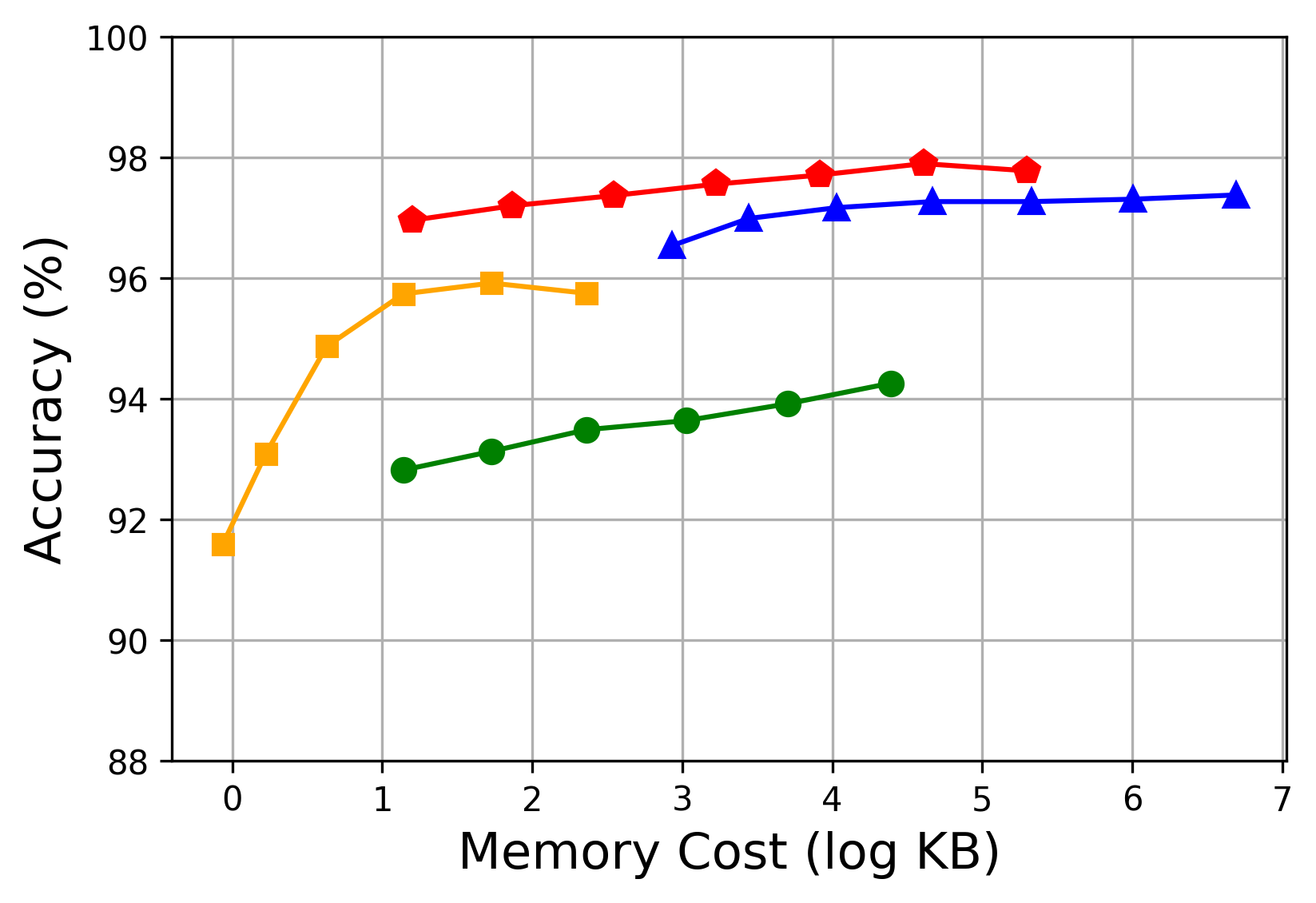}}
\\
\vspace{0.1in}
\subfigure[ijcnn]{\includegraphics[width=2.2in]{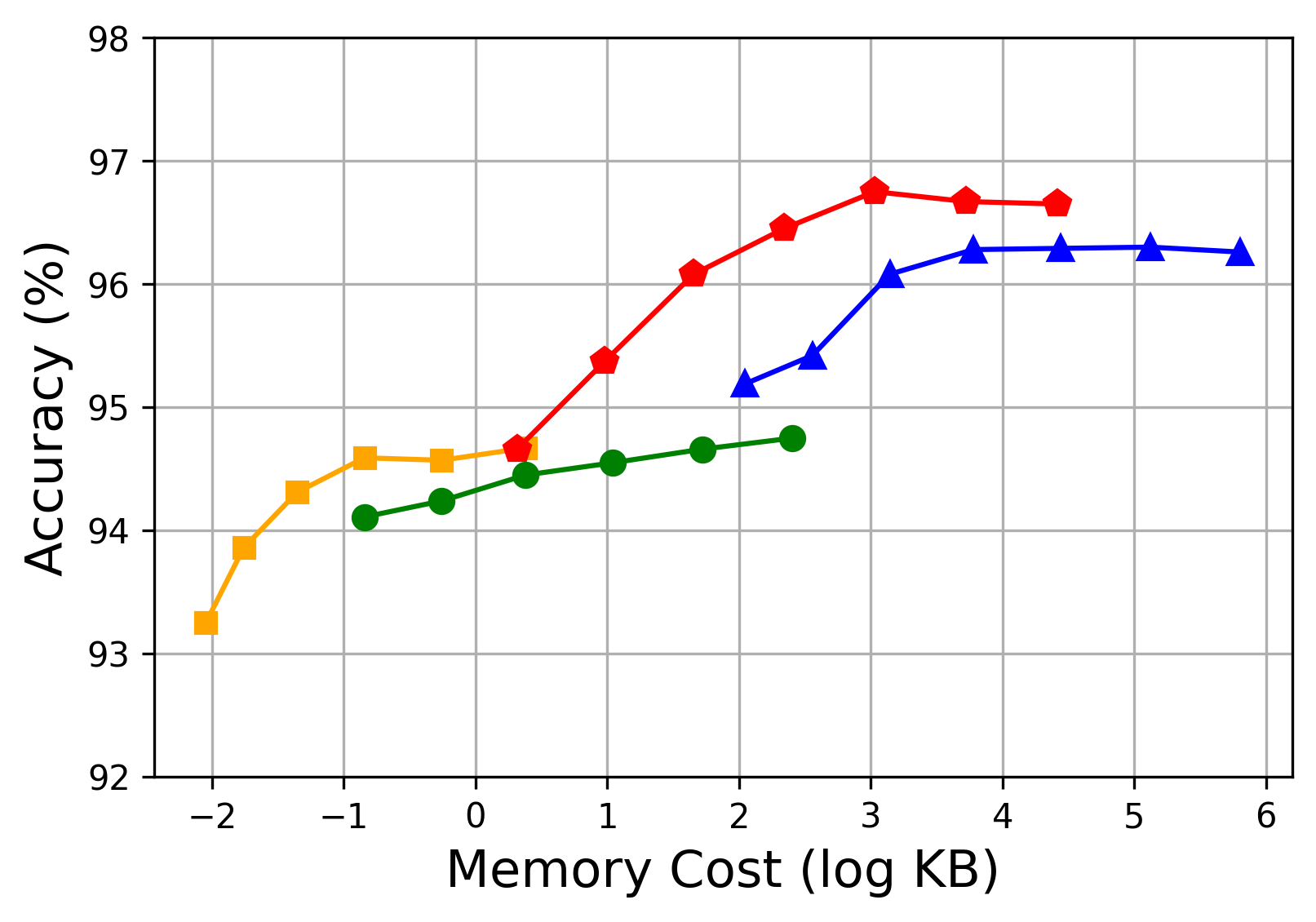}}
\hspace{0.05in}
\subfigure[webspam]{\includegraphics[width=2.2in]{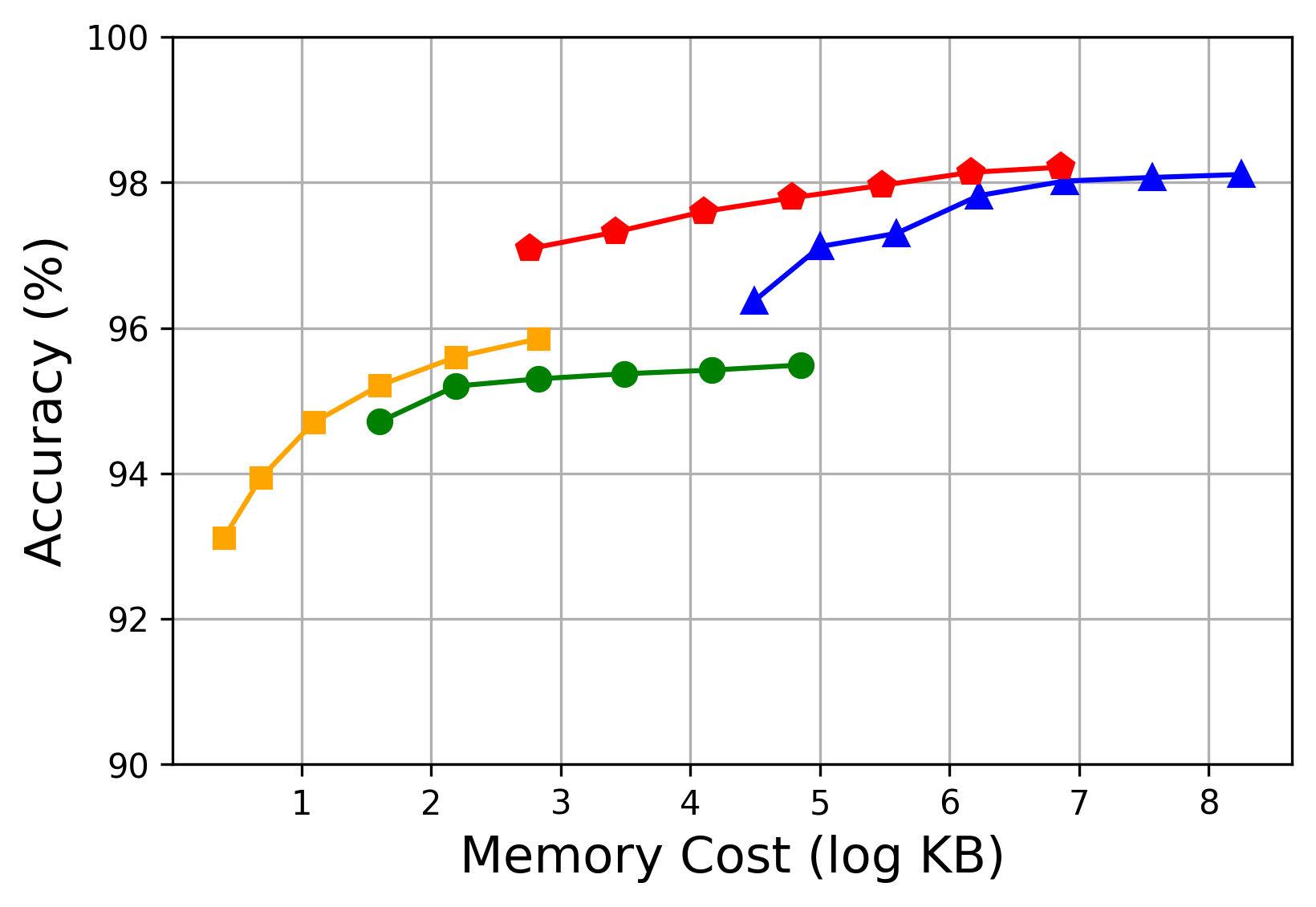}}
\end{center}
\caption{Memory cost vs. test accuracy of different FM models}
\label{fig:memory vs. accuracy}
\end{figure*}

\subsection{Experiment Results.}
The accuracy, prediction time and memory cost of all six algorithms are reported in Table \ref{table:evaluation_results}. Note that the memory cost of model parameters is normalized by the memory cost of standard FM. For example, 10x means the memory cost is 10 times larger than standard FM.
The best accuracy for each dataset is in bold. The second best accuracy is in italic. 

\textbf{Accuracy}. As shown in Table \ref{table:evaluation_results}, both SEFM and our proposed method Binarized FM achieve higher accuracy than Liblinear, FM and DFM on all eight datasets. This is due to the fact that the expressiveness capacities of Liblinear, FM and DFM are smaller than SEFM and our proposed binarized FM. Overall, the linear classifier (i.e., Liblinear) gets the lowest accuracy, while FM and DFM get higher accuracy than the linear classifier. DFM is less accurate than FM because of binarizing of feature interaction model parameters. Our proposed method Binarized FM achieves comparable results with SEFM with much lower memory cost.  

\textbf{Memory cost.} With regard to memory cost, Liblinear and DFM have much lower memory cost than FM, as Liblinear is a linear classifier being linear in the parameters and DFM binarize the 2-dimensional parameter $\mathbf{V}$ of FM. As for our proposed method Binarized FM, which binarizes both the parameter $\mathbf{w}$ and $\mathbf{V}$ and transforms features by one-hot encoding, it has relatively lower memory cost on {\sf circles}, {\sf moons}, {\sf pendigits}, {\sf ijcnn} and {\sf webspam} than FM and slightly higher memory costs on the other datasets. The memory costs of SEFM and NFM are much higher than FM and our proposed method on all of datasets. 

We visually present the variation between memory cost and accuracy among four methods--FM, SEFM, DFM and Binarized FM on six different datasets as shown in Figure \ref{fig:memory vs. accuracy}. It is clearly discovered that our proposed method Binarized FM can achieve higher accuracy than FM and DFM. With the same memory cost, Binarized FM has predominantly higher accuracy than FM.  SEFM has the high accuracy and the highest memory cost, while the memory cost of our proposed method Binarized FM is obviously lower and its accuracy approaches that of SEFM.

\textbf{The effect of scaling parameter $\alpha$ and $\beta$.}
The accuracy and prediction time of our algorithms with and without scaling parameter are  reported in Table \ref{table3}. After adding the scaling parameter $\alpha$ and $\beta$, our method achieves higher accuracy on most of datasets than without using them, especially on {\sf banana}, {\sf segment}, {\sf pendigits}, {\sf ijcnn} and {\sf webspam}.

\begin{table}[ht!]
    \caption{The effect of scaling parameter $\alpha$ and $\beta$ 
        }
        \begin{minipage}[b]{1\linewidth}\centering\normalsize
            \begin{tabular}{llcc}
            \toprule
            Dataset & Performance & Without & With \\
             $n$/$d$/\#class & & scaling & scaling \\
              & & parameter & parameter \\
            \midrule
            {\sf circles} & accuracy(\%) & 99.97$\pm$0.06 &  99.95$\pm$0.06 \\
            5,000/2/2 & prediction time  &  1.5ms   & 1.6ms \\
            \hline
            {\sf moons} & accuracy(\%) & 99.99$\pm$0.00 &  99.99$\pm$0.00 \\
            5,000/2/2 & prediction time  & 1.2ms   & 1.2ms  \\
           \hline
            {\sf banana} & accuracy(\%) & 87.01$\pm$0.60  & 89.25$\pm$0.48 \\
             5,300/2/2 & prediction time  & 1.5ms &  1.8ms   \\
            \hline
            {\sf breast-cancer} & accuracy(\%) & 96.00$\pm$1.06 & 96.20$\pm$0.64 \\
             683/10/2 & prediction time  & 0.8ms &  0.6ms  \\
            \hline
            {\sf segment} & accuracy(\%) & 92.29$\pm$0.51  & 94.75$\pm$0.82 \\
            2,310/19/7 & prediction time  & 17.5ms & 16.4ms \\
            \hline
            {\sf pendigits} & accuracy(\%) & 95.90$\pm$0.39  & 97.56$\pm$0.16 \\   
           10,992/16/10 & prediction time  & 1.5273s &  1.7003s   \\
            \hline
            {\sf ijcnn}   & accuracy(\%) & 95.13$\pm$0.20  & 96.45$\pm$0.09 \\
             49,990/22/2 & prediction time  & 0.2037s & 0.2292s \\
            \hline
            {\sf webspam} & accuracy(\%) & 96.71$\pm$0.08  & 97.79$\pm$0.08  \\
            350,000/254/2 & prediction time  & 7.1381s & 7.0044s   \\
            \bottomrule
            \end{tabular}
        \end{minipage}
        
        \label{table3}
\end{table}

\begin{figure}[H]
\begin{center}
\begin{tabular}{c}
\subfigure[FM on {\sf moons}]{\includegraphics[width=2.1in]{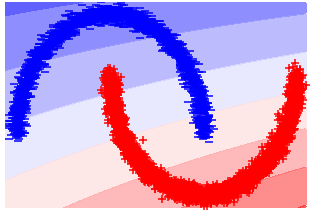}}
\hspace{.06in}
\subfigure[SEFM on {\sf moons}]{\includegraphics[width=2.1in]{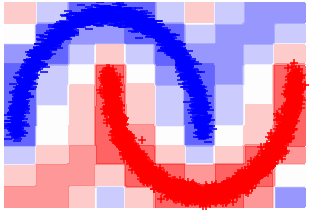}}\\
\subfigure[DFM on {\sf moons}]{\includegraphics[width=2.1in]{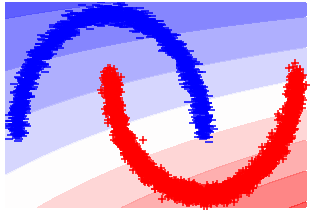}}
\hspace{.05in}
\subfigure[Binarized FM on {\sf moons}]{\includegraphics[width=2.1in]{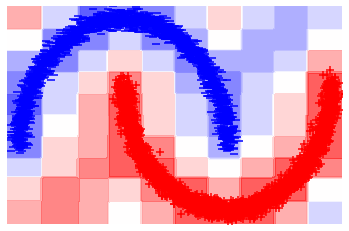}}\\
\subfigure[FM on {\sf banana}]{\includegraphics[width=2.1in]{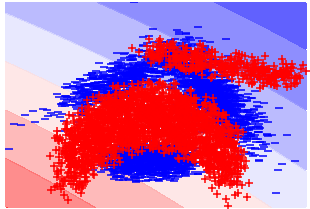}}
\hspace{.06in}
\subfigure[SEFM on {\sf banana}]{\includegraphics[width=2.1in]{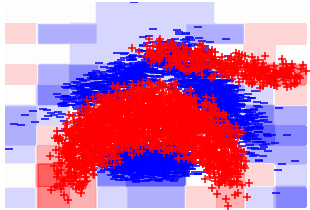}}\\
\subfigure[DFM on {\sf banana}]{\includegraphics[width=2.1in]{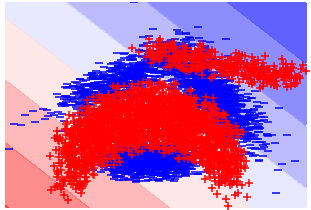}}
\hspace{.05in}
\subfigure[Binarized FM on {\sf banana}]{\includegraphics[width=2.1in]{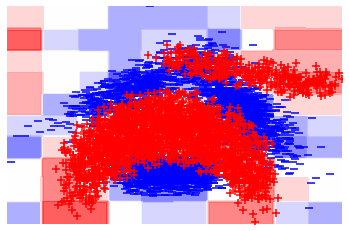}}
\end{tabular}
\end{center}
\caption{Decision boundaries of FM, SEFM and our method on {\sf moons} and {\sf banana} datasets}
\label{fig:moons_banana_decision_boundaries}
\end{figure}

\textbf{Illustration of Decision Boundaries on Synthetic Datasets.} We first illustrate that our proposed method can produce highly nonlinear decision boundary as SEFM on the synthetic datasets. We show the decision boundaries of FM, SEFM, DFM and our proposed binarized FM on {\sf moons} and {\sf banana} datasets in Figure \ref{fig:moons_banana_decision_boundaries}. As shown in this figure, both FM and DFM can not capture the highly nonlinear pattern on these two datasets since they only used polynomial expansion on the original data to model the nonlinear relationship. SEFM and our proposed binarized FM can produce complex piecewise axis perpendicular decision boundary.

\begin{figure}[htb]
\begin{center}
\begin{tabular}{c}
\subfigure[banana]{\includegraphics[width=3in]{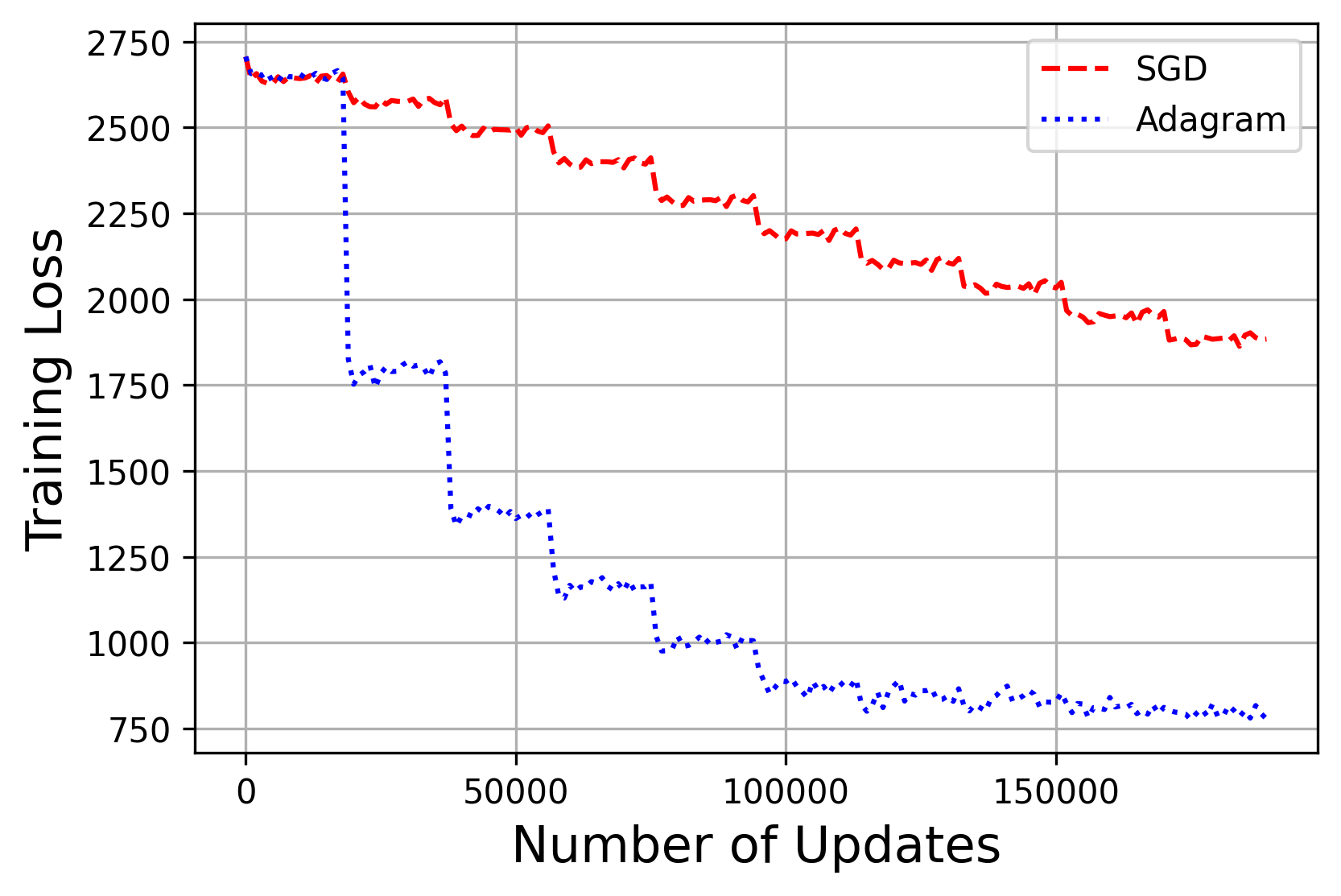}}\\
\subfigure[ijcnn]{\includegraphics[width=3in]{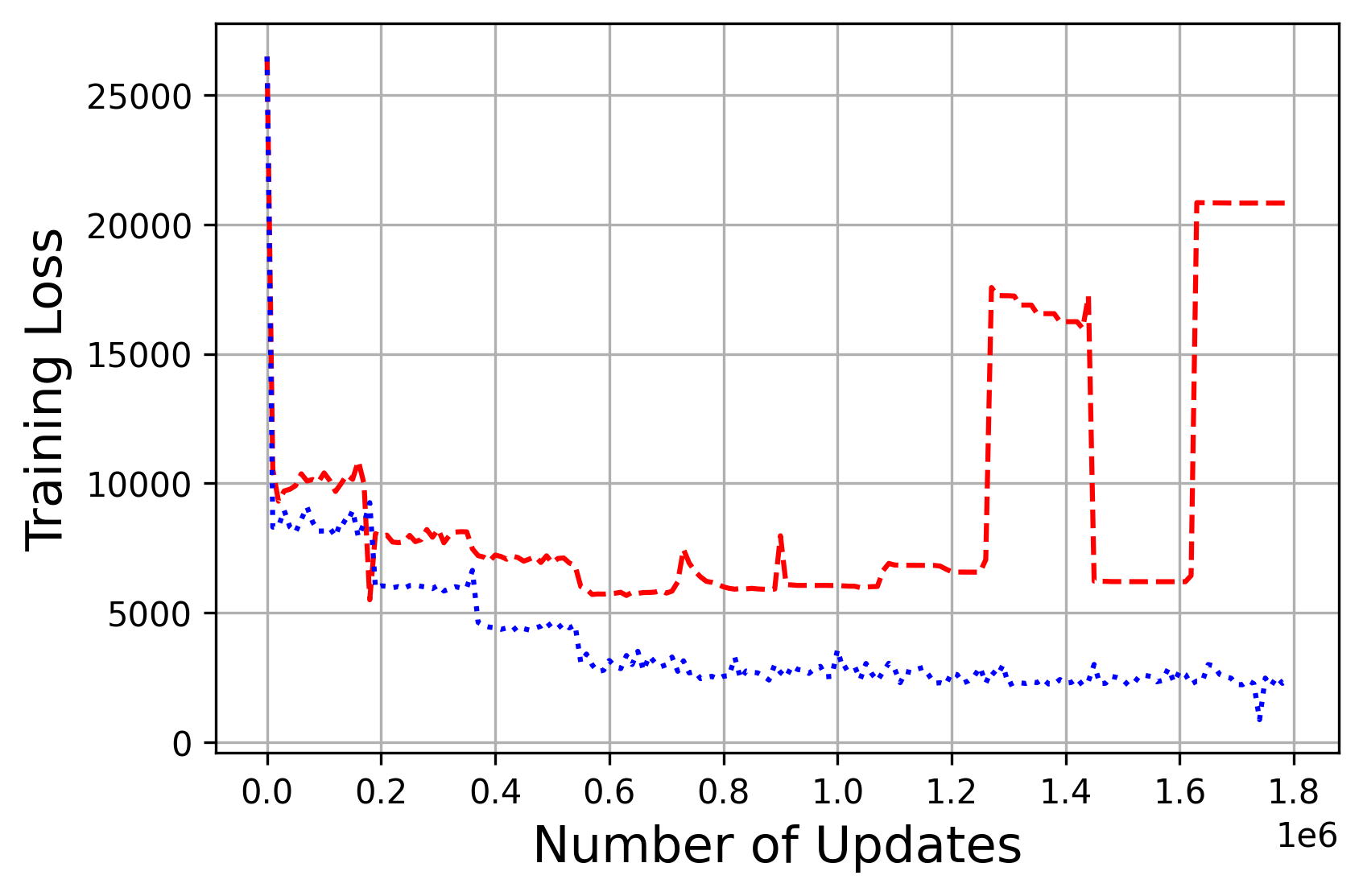}}
\end{tabular}
\end{center}
\caption{Comparison of the Convergence of Adagrad and SGD}
\label{fig:training loss vs. iteration}
\end{figure}

\textbf{Comparison of the Convergence of Adagrad and SGD in Binarized FM.}
In addition to demonstrating the performance of our proposed method, we illustrate our training process by presenting the variation of training loss during iteration to prove our optimization method--Adagrad has more efficient performance than SGD. As shown in Figure \ref{fig:training loss vs. iteration}, Adagrad reaches the lower training loss fast and gradually tend to converge.

\newpage

\section{Conclusion}
In this paper, we propose a novel method Binarized FM that can reduce the memory cost of SEFM by binarizing the model parameters while maintaining comparable accuracy as SEFM. Since directly learning the binary model parameters is an NP-hard problem, we propose to use STE with Adagrad to efficiently learn the binary model parameters. 
We also provide a detailed analysis of our proposed method's space and computational complexities for model inference and compare it with other competing methods. Our experimental results on eight datasets have clearly shown that our proposed method can get much better accuracy than FM while having similar memory cost 
and can achieve comparable accuracy with SEFM but with much less memory cost. We also demonstrate that learning model parameters by STE with Adagrad can converge very fast during the iteration of training the input dataset.
\bibliographystyle{apalike}
\bibliography{myRef}   

\end{document}